\documentclass[10pt,twocolumn,letterpaper]{article}

\usepackage{iccv}
\usepackage{times}
\usepackage{epsfig}
\usepackage{graphicx}
\usepackage{amsmath}
\usepackage{amssymb}
\usepackage[numbers]{natbib}
\usepackage{multirow}
\usepackage{tablefootnote}
\usepackage{adjustbox}
\usepackage[table,xcdraw]{xcolor}

 \makeatletter
\@namedef{ver@everyshi.sty}{}
\makeatother

\usepackage{bm}
\usepackage{booktabs}
\usepackage{multirow}
\usepackage{array}
\usepackage{pgfplots}
\usepackage{fontawesome5}
\newcommand{\cirbd}{\mathrel{\text{\faDotCircle[regular]}}}
\usepackage{dsfont}
\usepackage[rightcaption]{sidecap}
\usepackage{wrapfig}
\usepackage{caption}
\pgfplotsset{width=8cm,compat=1.9}

\iccvfinalcopy 

\begin{document}

\title{Unsupervised Domain Adaptation for Training Event-Based Networks Using Contrastive Learning and Uncorrelated Conditioning}

\author{Dayuan Jian\\
University of Southern California\\
{\tt\small djian@usc.edu}
\and
Mohammad Rostami 
\\University of Southern California\\{\tt\small rostamim@usc.edu}
}

\maketitle

\begin{abstract}
   Event-based cameras offer reliable measurements for preforming computer vision tasks in high-dynamic range environments and during fast motion maneuvers. However, adopting deep learning in event-based vision faces the challenge of annotated data scarcity due to recency of event cameras. Transferring the knowledge that can be obtained from conventional camera annotated data offers a practical solution to this challenge. We develop an unsupervised domain adaptation algorithm for training a deep network for event-based data image classification using contrastive learning and uncorrelated conditioning of data. Our solution outperforms the existing algorithms for this purpose.
\end{abstract}

\section{Introduction}
\label{sec:intro}
Inspired by biological visual systems, event-based cameras are designed to measure image pixels independently and asynchronously through recording changes in brightness, compared to  a reference brightness level~\cite{lichtsteiner2008128}. 
A stream of events is measured which encodes the location, time, and polarity of the intensity changes. Compared to conventional frame-based cameras, they offer high dynamic range, high temporal resolution, and low latency which make them ideal imaging devices for preforming computer vision (CV) tasks that involve high dynamic range and fast movements. Despite this promising prospect, adopting artificial intelligence and particularly deep learning to automate CV tasks in this domain is challenging. The primary reason is that training deep neural networks relies on manually annotated large datasets. However, annotated event-data is scarce due to the recency of these cameras: event-based data represent only $3.14\%$ of existing vision data~\cite{hu2020learning}.

Unsupervised domain adaptation (UDA)~\cite{ganin2015unsupervised} is a learning framework to train a model for a target domain with unannotated data through transferring knowledge from a secondary source domain with annotated data. The core idea in UDA is to reduce the distributional domain gap by mapping data from both domains into a shared domain-agnostic embedding space~\cite{baktashmotlagh2013unsupervised,bousmalis2017unsupervised,tzeng2017adversarial,pan2019transferrable,rostami2021transfer,stan2022domain,rostami2019deep}. Although event-based and frame-based cameras are significantly different, the intrinsic relationship between their outputs through the real-world relates the information content of the outputs considerably.  As a result, a frame-based annotated dataset can be served as a source domain for an event-based domain as the target domain in the context of a UDA problem. Even if the related frame-based data is unannotated, annotating frame-based data is much simpler.

The above possibility has led to a line of UDA algorithms for event-based CV tasks. Initial works in this area consider datasets with paired events and frames, i.e., the pixel-level recordings describe the same input~\cite{hu2020learning,zhu2021eventgan}. Despite being effective, these methods are highly limited because they consider that both measurement types are preformed simultaneously. Recently, a few UDA algorithms have been developed for unpaired event-based data~\cite{rebecq2019high,Messikommer20ral}.
An effective idea is to use the idea of translation in UDA~\cite{murez2018image} and generate events for corresponding video data~\cite{gehrig2020video,hu2021v2e}.
An alternative idea is  to generate event-based annotated data synthetically~\cite{rebecq2018esim} and use it as the source domain~\cite{planamente2021da4event}.

 We develop a new UDA algorithm for event-based data using unpaired frame-based data. Our contribution is two-fold.
 First, we use contrastive learning in combination with data augmentation to improve the model generalizability. The idea is to project different augmentation types of one object to a latent representation and train an encoder to maintain identities for all augmentation types. Second, we introduce an uncorrelated conditioning loss term. The new loss term is used to regularize  the model with additional information: the latent vector representation of an object under event cameras should be uncorrelated to how the object looks like under event cameras. Experiments  show that our method leads to learning a better representation of events in an unsupervised fashion. Our results indicate   2.0\% on Caltech101 $\rightarrow$ N-Caltech101 and 3.7\% on CIFA10 $\rightarrow$ CIFA10-DVS performance improvements over state-of-the-art.

\section{Related Work}
\label{sec:Related}
\paragraph{Unsupervised Domain Adaptation for Frame-Based Tasks:}
UDA has been explored extensively in CV when both domains are frame-based measurements from the same modality. For a thorough and extensive survey, please refer to the work by Wilson and Cook ~\cite{wilson2020survey}. The core idea for UDA in frame-based tasks is to train an encoder that is shared across the two domains and  map the training data points from both domains into the shared domain-agnostic latent embedding space, i.e., modeled as the encoder  output space. As a result, if we use the source domain labeled data and train a classifier that receives its input from the embedding space, it will generalize well on the target domain due to the distributional alignment of both domains in the shared embedding space.
The domain-agnostic embedding has been  learned primarily using either  generative adversarial learning~\cite{he2016deep,sankaranarayanan2018generate,pei2018multi,zhang2019domain,long2018conditional} or by minimizing a suitable distribution alignment loss term~\cite{long2015learning,ganin2014unsupervised,long2017deep,kang2019contrastive,rostami2020sequential,rostami2021lifelong} in the UDA literature.
In the first approach for UDA, the generator subnetwork within the generative adversarial learning architecture~\cite{goodfellow2020generative} is served as the shared encoder  which is trained to compete against a discriminative subnetwork to learn a domain-agnostic, yet discriminating embedding space. As a result of this process,  the two distributions are aligned indirectly at the output of the generative subnetwork.
In the second approach, the distance between the empirical distributions of the two domains is measured at the output of the shared encoder, e.g., using a suitable probability distribution metric such as KL-divergence or sliced optimal transport~\cite{stan2021unsupervised,rostami2022increasing}, and the encoder is  then trained such that   that  distance is minimized directly.

\paragraph{Unsupervised Domain Adaptation for Event-Based Tasks:} despite the challenge of annotated data scarcity for event-based data, only recently some works have tried to benefit from UDA framework to address tasks with unannotated event-based data.  The reason is that when the source and the target domain have different   modalities, e.g., frame-based and event-based data, the distribution gap between the two domains is significantly larger. For this reason, using a shared encoder may not be applicable  for aligning the distributions.
However, we can benefit from the special case of   image-to-image translation~\cite{murez2018image} from the UDA literature. The core idea of image-to-image translation is to map images from either of the two domains to the other domain using a generative model. As a  result,  we can match the distributions directly at the input image space of one of the domains. For example, if we translate the target domain images to the source domain, we can directly use the source-trained classifier.
For this reason, a group of exiting works translate event data frame-based images using adversarial training~\cite{wang2019event,wang2020eventsr,choi2020learning}.
A major limitation for these works is that we require  paired image-event data that is
measured on the same scene to learn cross-domain correspondences. This is a strong limitation because it means that we should measure and generate paired data to build new datasets which in some applications is not even feasible, e.g., when event and frame sensors are used in different lighting conditions. This requirement has been relaxed by 
separating domain-agnostic features, e.g. geometric structures, from
the domain-specific feature, e.g., texture~\cite{zhang2020learning}.
Another  existing approach to tackle this challenge is to benefit from the idea of video-to-event translation using  
generative models with proper architecture. 
The goal is to convert frame-based video into synthetic event-based data to train the model directly in the event space.
To this end, we can either rely on model-based translation~\cite{rebecq2019high,gehrig2020video,hu2021v2e} or data-driven translation~\cite{zhu2021eventgan}.
These methods
are still limited because they can only translate videos to events, thus making them inapplicable  to image datasets which form the majority of existing datasets.
 The work
most similar to our work  splits the embedding space into shared and sensor-specific features. An event generation model is then used to
align both domains~\cite{Messikommer20ral}.
In our work, we benefit from contrastive learning to improve model generalizability and define a novel loss term that makes representations of objects uncorrelated to their even-based measurements. Our experiments demonstrate that our contributions can help to achieve the performance level of supervised learning using UDA, relaxing the need for data annotation in some cases. 

\section{Domain Adaptation from Frames to Events}
\label{sec:Description}
\begin{figure}[h]
    \centering       
    \includegraphics[width=7.3cm, height=2.2cm]{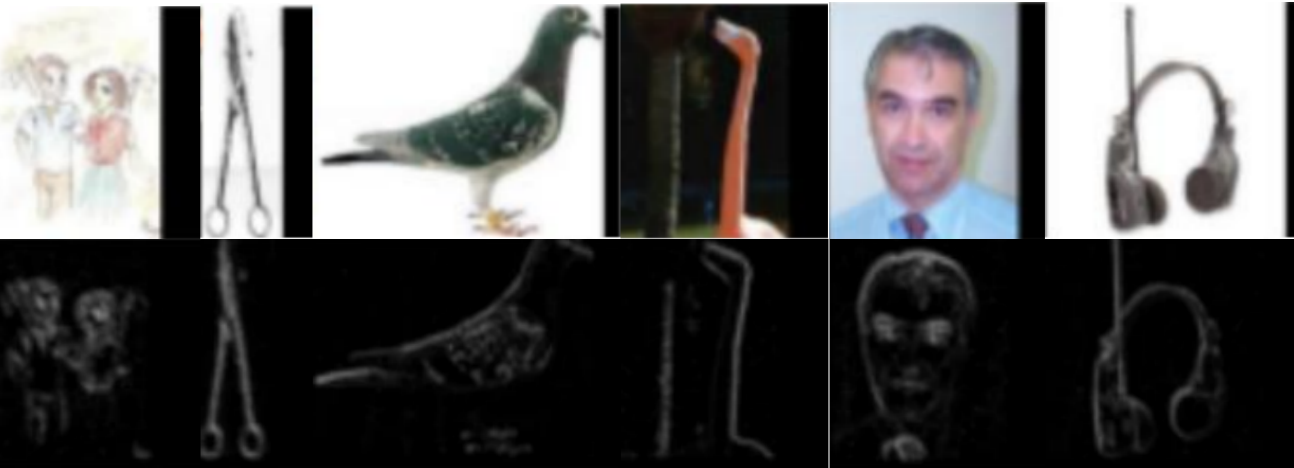}
\vspace{-2mm}
 \caption{Randomly drawn frame (row 1, Caltech101\cite{https://doi.org/10.48550/arxiv.1604.01518}) and event (row 2, N-Caltech101\cite{https://doi.org/10.48550/arxiv.1507.07629})  pairs }
    \label{fig1}
    \vspace{-2mm}
\end{figure}

Our goal is to train a model on labeled frames (RGB images) and then use  the trained network on events data, without requiring any labeled events. More importantly,  we assume that paired frames and event data is not accessible for our task of interest.
Figure~\ref{fig1} presents samples of paired frame and event data for two common event-based datasets. As it can be seen, the pairs are  significantly more different than benchmarks that are used in the frame-based  UDA literature, e.g., Office-Home~\cite{venkateswara2017deep} or DomainNet~\cite{peng2019moment} dataset. 
For this reason, UDA is a challenging problem for tasks that have event-based and frame-based data as their target and source domains. 
Note, however, event and frame-based cameras
capture scene in physical work that share strong information overlap which makes knowledge transfer across these two domain feasible. The idea that we will explore is to align the content representations obtained from events to the ones from frames so that we can use the frame classifier.

More specifically, consider that we have access to unannotated event data instance $\bm{y}_e$ along with annotated frame-based image data  $\bm{y}_f$ that are drawn from the event domain $\mathcal{Y}_e$ and the frame domain $\mathcal{Y}_f$, respectively. Our goal is to map $\bm{y}_e$ and $\bm{y}_f$  into a shared domain-agnostic embedding space $\mathcal{Z}$. We refer to this space as content space because the goal is to learn it such that shared information content can be represented in the shared embedding space.
Let   $E_f(\cdot), E_{e,cont}(\cdot)$ be two suitable deep encoders that project frames to the content feature space and events to the content feature space.
We use two independent encoders due to the large domain gap between these two domains. Also, let  $C(\cdot)$ denote a classifier network that
maps the extracted content features to the label space. Since frames and their labels are provided, we can easily use standard supervised learning to train the neural network $C(E_f(\bm{y}_f))$ using empirical risk minimization for predicting frame labels.  

Our ultimate goal is to train a model for event domain. However,
the above approach is not applicable to the event  domain because the labels are missing.
The idea that we would like to use is to align event content representation $E_{e,cont}(\bm{y}_e)$ with frame content representation $E_f(\bm{y}_f)$. If we have access to paired event-frame data, we can just simply use a point-wise loss function to align the data points, but we are considering a more practical setting, where we don't have frame-event pairs. 
We develop an algorithm for this purpose. Upon aligning the two domains, we can classify the even data using the model $C(E_{e,cont}(\bm{y}_e))$.

\section{Proposed Method}
\label{sec:Method}
\begin{figure}[h]
    \centering
    \includegraphics[width=6.5cm, height=7 cm]{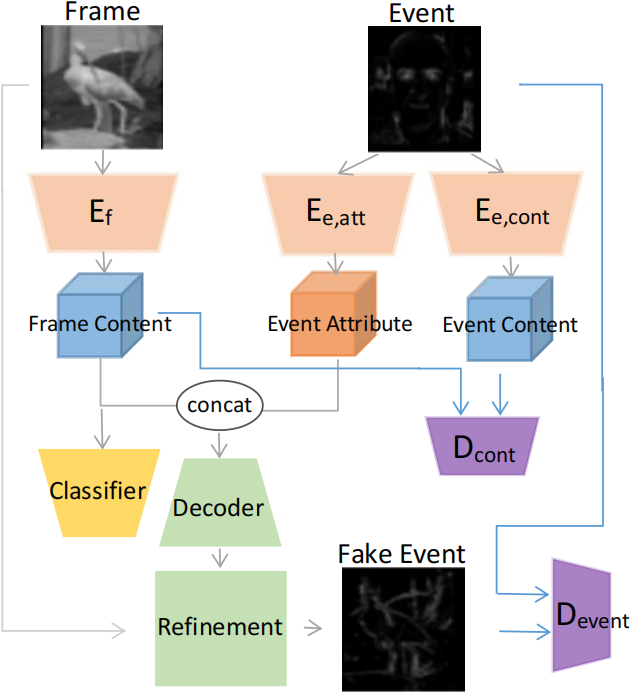}
    \vspace{-2mm}
    \caption{\small Architecture for our method during training.
    During testing, only event content encoder and classifier are needed.}
    \label{fig2}
    \vspace{-2mm}
\end{figure}

The architecture of our framework is presented in Figure \ref{fig2}.
Our core idea is to benefit from adversarial training and  generate fake events from content features of frame images such that they look like real events. As a byproduct of generative adversarial learning, the content features from both domain are aligned indirectly.

 As shown in Figure \ref{fig2}, an image encoder and an event encoder generate  features that represent the content inside  images and  events, respectively. The output space of these two encoders share the same dimension. The content features are then fed into the classifier to predict labels.  The goal is to train these encoders such that event content representations have the same distribution of the frame content representations in $\mathcal{Z}$.  Additionally, we have a second event encoder $E_{e,att}(\cdot)$ that generates feature attributes that intuitively correspond to how objects look like under an event camera.  
 The frame content and the event attribute are sent to a decoder to generate a fake event. The fake event and the frame are sent to a refinement network to get a  fake event with a higher quality. This fake event should contain the content inside the frame and present the content under an event camera.
Clearly, if the fake event looks realistic, it means that the shared content encodes shared information from events and frames.

To close the adversarial learning architecture, we also use  a content discriminator network $D_{cont}(\cdot)$ 
to discriminate whether the content representation is from an event or an image. This network is trained against both encoders to learn  domain-agnostic content features. To enforce more constraints, we also rely on an idea similar to cycleGAN \cite{CycleGAN2017}. 
The generated fake event is fed into the two event encoders and its event attribute and event content are computed. They are compared with the ones which generated the fake event using \(L_1\) loss.
To generate  fake events that are realistic, we consider a second discriminator network  $D_{event}(\cdot)$ 
used to discriminate real and fake events. Since the fake event is generated from content features which we know its label, the content feature of the fake event is also sent to the classifier for further training. The choice for encoders and the classifier is a design choice. We can also use similar architectures for the three encoders for simplicity. To implement the above constraints, we  define the following loss terms and solve the underlying cumulative optimization problem according to the architecture presented in Figure~\ref{fig2}:
\begin{itemize}
 \item     \textbf{Classification loss on frames, $\mathcal{L}_{cls,frame}$}, assumed to be a
Cross Entropy loss function.
 \item  \textbf{Classification loss on fake events, $\mathcal{L}_{cls,fake}$},  Cross Entropy loss  on classification of fake events.
 \item  \textbf{Decoder output loss, $\mathcal{L}_{decoder}$}: it consists of two terms. The first term  approximates the event version of a frame image according $\ell_1$ loss by the work of Gallego, Guillermo et al\cite{Gallego2015EventbasedCP}.  The second term enforces  cycle consistency \cite{CycleGAN2017} for improvement. The decoded outputs of event attribute of fake events and frame contents are enforced to be  close to the decoded output of event attributed of real event and frame content.  
\item \textbf{Cycle loss on feature representations of contents and attributes:} $\mathcal{L}_{cyc,cont}$ and $\mathcal{L}_{cyc,att}$ 
These are $\ell_1$ losses imposed on the features used to generate the fake events and the features extracted from them:
\begin{equation}
\small
\begin{split}
&\mathcal{L}_{cyc,cont} = ||E_f(\bm{y}_f)-E_{e,cont}(fake)||_1\\&\mathcal{L}_{cyc,att} = ||E_{e,att}(\bm{y}_e)-E_{e,att}(fake)||_1
\end{split}
\end{equation}
\item \textbf{Content Discriminator:} There is a discriminator network to differentiate frame content and event content. The purpose of this discriminator is to make frame content features and event content features have the same distribution. Denoted $g(\cdot)$ = \(\log(sigmoid(\cdot))\) and $h(\cdot)$ = \(\log(1-sigmoid(\cdot))\). We use relativistic average discriminator\cite{jolicoeurmartineau2018relativistic} loss as follows:
\begin{equation}
\small
\begin{split}
&\mathcal{L}_{dis,cont} =   \frac{1}{N}\sum 
\\&\{\bm{g}[D_{cont}(E_f(\bm{y}_f)) -\frac{1}{N}\sum D_{cont}(E_{e,cont}(\bm{y}_e))] +\\& \bm{h}[D_{cont}(E_{e,cont}(\bm{y}_e))-\frac{1}{N}\sum D_{cont}(E_f(\bm{y}_f))]\}
\end{split}
\end{equation}
\item Generative Loss for encoder $E_f, E_{e,cont}$  as follows: 
\begin{equation}
\small
\begin{split}
&\mathcal{L}_{encoder,cont} =   \frac{1}{N}\sum\\&\{\bm{h}[D_{cont}(E_f(\bm{y}_f)) -\frac{1}{N}\sum D_{cont}(E_{e,cont}(\bm{y}_e))] + \\&\bm{g}[D_{cont}(E_{e,cont}(\bm{y}_e))-\frac{1}{N}\sum D_{cont}(E_f(\bm{y}_f))]\}
\end{split}
\end{equation}

\item \textbf{Event Discriminator loss:} There is a discriminator network to differentiate fake events and real events. The purpose of this discriminator network is to make fake events and real events have the same distribution. We use a relativistic average discriminator\cite{jolicoeurmartineau2018relativistic}:
\begin{equation}
\small
\begin{split}
&\mathcal{L}_{dis,e} = 
 \frac{1}{N}\sum \{\bm{g}[D_{e}(\bm{y}_e) -\frac{1}{N}\sum D_{e}(fake)]\\& + \bm{h}[D_{e}(fake)-\frac{1}{N}\sum D_{e}(\bm{y}_e)]\}
\end{split}
\end{equation}
\item Similarly, the generator loss is: 
\begin{equation}
\small
\begin{split}
&\mathcal{L}_{gen,e} =\frac{1}{N}\sum  \{\bm{h}[D_{e}(\bm{y}_e) -\frac{1}{N}\sum D_{e}(fake)] + \\&\bm{g}[D_{e}(fake)-\frac{1}{N}\sum D_{e}(\bm{y}_e)]\}
\end{split}
\end{equation}

\item \textbf{Orthogonal Normalization:} $\mathcal{L}_{orth}$, introduced in BigGAN \cite{brock2019large}, and is imposed on   discriminators. This regularization term is crucial to stabilize the training.
\begin{equation}
\small
\begin{split}
L_{orth} = \beta ||W^TW\cirbd(\mathds{1}-I)||^2_F,
\end{split}
\end{equation}
where \(W\) is the weight matrix, \(\mathds{1}\) is the matrix where all elements are 1, \(I\) is the identity matrix, \(\cirbd\) is element-wise multiplication, and denotes $\|\cdot\|_F$ Frobenius norm.

\end{itemize}

We can minimize sum of all or a reasonable subset of above loss terms for domain alignment~\cite{planamente2021da4event}.  In this work, we propose two novel ideas based on contrastive learning and uncorrelated conditioning for further improvement.

\begin{itemize}
    \item   \textbf{Constrastive Learning:} Figure ~\ref{fig3} visualizes our pipeline for computing this term.

    \begin{figure}[h]
    \centering
    \vspace{-3mm}
    \includegraphics[width=7cm, height=5.8cm]{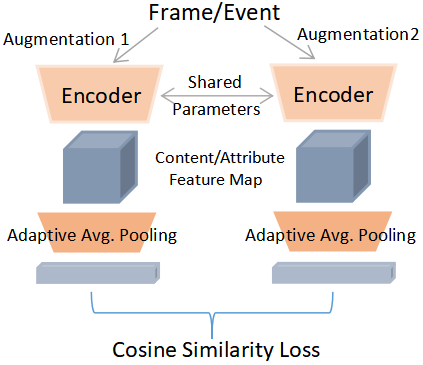}
    \vspace{-3mm}
    \caption{Contrastive learning loss, used for all 3 encoders. }
    \label{fig3}
\end{figure}
    
    The goal is to use augmentation and generate different  versions of an image and enforce them to share the same feature representations in the latent space \cite{https://doi.org/10.48550/arxiv.2006.07733,chen2020big,chen2020mocov2}. Contrastive learning loss serves as a constraint to help convergence to a better solution. While constrastive learning algorithms usually use negative samples as well, we only consider positive samples. Denote \(f(\cdot)\) as encoder plus average pooling. Then \(f(\bm{y})\) and \(f(\bm{y}')\) will be two feature vector representations of a single input \(\bm{y}\), produced using augmentation. We then enforce the following hard equality constraint:
\begin{equation}
\frac{\langle f(\bm{y}),f(\bm{y}')\rangle}{||f(\bm{y})||_2\cdot||f(\bm{y}')||_2} = 1, 
\end{equation}
where $\langle\cdot,\cdot\rangle$ denotes vector inner product and $||\cdot||_2$ is  $l_2$-norm. We enforce this constraint on outputs of all the three encoders for data from both domains.

\begin{figure}[h]
    \centering
    \includegraphics[width=7cm, height=5.6cm]{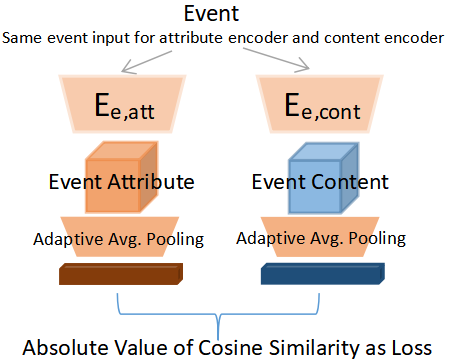}
    \vspace{-3mm}
    \caption{Uncorrelated Conditioning}
    \label{fig4}
    \vspace{-4mm}
\end{figure}

\item  \textbf{Uncorrelated Conditioning:} to the best of our knowledge, imposing this constraint on two different features which extracted from the same input is novel and unprecedented. In most tasks, we mostly care about objects captured in an image. The intuition behind this loss is to benefit the full information content of an image.   In addition to objects in an image, e.g., a pedestrian,   we can deduce other information from the totality of an image. For example, whether the image is blurred or not, whether the image is taken at a day or  a night, whether the camera is  high-resolution or  low-resolution, whether it is taken by a frame camera or an event camera, etc. These are also features we can extract from images. However, these properties are uncorrelated to each other, i.e., whether the image is blurred or not can tell nothing about whether there is a pedestrian in the image, whether the image is taken at a day or a night can tell nothing about the quality of the used.  
Our idea is to train the model not only to extract features about objects  inside an image, but  also  to extract domain features. In our method, we need to extract features from event data because objects captures by event cameras look  different from the same objects captured by a frame cameras. This is the major reason that we produce fake or translated event  images. Therefore, we need to impose uncorrelated constraint on the object features and event camera features, e.g., whether an object is a pedestrian or something else in the frame image can tell nothing about how the corresponding event image would look like. Similarly,  how an event image looks like would reveal no information about the content of the event image. We enforce this intuition using the uncorrelated conditioning constraint:
\begin{equation}
\frac{\langle f_{e,a}(\bm{y}_e),f_{e,c}(\bm{y}_e)\rangle}{||f_{e,a}(\bm{y}_e)||_2\cdot||f_{e,c}(\bm{y}_e)||_2} = 0,
\label{eq8}
\end{equation}
where \(f_{e,a}(\bm{y}_e)\) and \(f_{e,c}(\bm{y}_e)\) are the two feature vector representations of attribute and content, respectively, for the same event \(\bm{y}_e\). 
The constraint in Eq.~\ref{eq8} helps to decorrelate 
attribute and content for events.
\end{itemize}
We further denote \(f_{fr}(\bm{y}_f)\) as the content vector for frame \(\bm{y}_f\).
We add the above two terms to the optimization objective function as two additional regularizing loss terms:

\begin{equation} 
\begin{split}
&\min \sum_{y_e,y_f}  
-\frac{\lambda_1\cdot\langle f_{e,a}(\bm{y}_e),f_{e,a}(\bm{y}_e')\rangle}{||f_{e,a}(\bm{y}_e)||_2\cdot||f_{e,a}(\bm{y}_e')||_2} \\&- 
\frac{\lambda_2\cdot\langle f_{e,c}(\bm{y}_e),f_{e,c}(\bm{y}_e')\rangle}{||f_{e,c}(\bm{y}_e)||_2\cdot||f_{e,c}(\bm{y}_e')||_2}  - \frac{\lambda_3\cdot\langle f_{fr}(\bm{y}_f),f_{fr}(\bm{y}_f')\rangle}{||f_{fr}(\bm{y}_f)||_2\cdot||f_{fr}(\bm{y}_f')||_2}\\&
+ \lambda_4\cdot|\frac{\langle f_{e,a}(\bm{y}_e),f_{e,c}(\bm{y}_e)\rangle}{||f_{e,a}(\bm{y}_e)||_2\cdot||f_{e,c}(\bm{y}_e)||_2}|,
\end{split}
\label{eq17}
\end{equation}
where \(\lambda_1\), \(\lambda_2\), \(\lambda_3\) and \(\lambda_4\)  are regularization parameters.

\section{Experimental Results}
\label{sec:Results}


\subsection{Experimental Setup}

\textbf{Datasets:} We evaluated our approach on  N-Caltech101\cite{https://doi.org/10.48550/arxiv.1507.07629} and CIFAR10-DVS\cite{lievent} event dataset.

\begin{itemize}
    \item  N-Caltech101\cite{https://doi.org/10.48550/arxiv.1507.07629} is an event classification dataset which is the event-based version of the Caltech101\cite{https://doi.org/10.48550/arxiv.1604.01518} dataset. Hence, we can define a UDA task to adapt a model from Caltech101 to N-Caltech101. 
    The N-Caltech101 dataset events are paired with the Caltech101 dataset RGB images. During training, the input RGB images and events are not paired and the label of events are   not provided either. 
    Both Caltech101 and N-Caltech101 are split into 45\% training set, 30\% validation set, and 25\% test set. Our test set set is exactly the same as previous works. We reduced the size of training set due to our very tight computing resources.

\item CIFAR10-DVS\cite{lievent} dataset is the event version of CIFAR10\cite{Krizhevsky2009LearningML}. There exists a huge gap between CIFAR10 and CIFAR10-DVS, i.e. CIFAR10 is a relatively easy dataset while CIFAR10-DVS is much more challenging. The large gap poses difficulty to perform UDA. Additionally, CIFAR10 and CIFAR10-DVS datasets are not paired. To our best knowledge, we are the first one to try UDA approach on CIFAR10-DVS. Following the literature, both CIFAR10 and CIFAR10-DVS are split into 5/1 training/testing splits.

\end{itemize}

\textbf{Data Augmentation for Contrastive Learning:}
To compute the constrastive learning loss, we randomly applied color jitter, Gaussian blur, random resize, random affine, random crop, and random flip on frame images to generate the augmented versions. In the events domain,  we used random event shift, random flip, random resize,  and random crop. These diverse transformations help improve the generalizability of our model and increasing model robustness with respect to domain shift in the input spaces.

\textbf{Architecture of Networks:} we used the first half of a ResNet18\cite{https://doi.org/10.48550/arxiv.1512.03385}, except tath the first Conv layer is changed to allow for different input sizes and max pooling layers are removed. The classifier is assumed to be the second half of ResNet18. Hence, we align the distributions in the middle layer of ResNet. Discriminator networks are simple CNNs.

\textbf{Optimization:}
R-Adam  optimizer~\cite{https://doi.org/10.48550/arxiv.1908.03265} is used with \begin{math}\beta_1 = 0\end{math}, \begin{math}\beta_2 = 0.999\end{math} with    \begin{math}lr = 10^{-4}\end{math}, and an exponential learning rate decay of 0.95. We train  on GeForce RTX 2080 Ti GPU. For Caltech101 and N-Caltech101, the batch size is 7 and training epoch is 150 which took 130 hours to run. For CIFAR10 and CIFAR10-DVS, the batch size is 21 and training epoch is 70 which took 36 hours to run.

\subsection{Performance Results}

We first report our performance on the two benchmarks and compare our results against existing state-of-the-art event-based UDA methods.
In our results,  We have also included results for a second group of supervised learning algorithms that benefit from transfer learning, but assume event labels are accessible   during training. 
All supervised methods on CIFAR10-DVS designed new models specific to events and used different way to represent events. We have also included a ``Baseline'' version of our approach which is trained using all loss functions, except contrastive learning and uncorrelated conditioning.  We refer to our method as Domain Adaptation for event data using Contrastive learning and  uncorrelated Conditioning  (DAEC$^2$) in our tables.

Tables \ref{tabl1} and \ref{tabl2} present our results.
As it can be seen from the tables, even a Baseline version of our approach achieves  3.8\% and a 3.1\% increased performances on N-Caltech101 and CIFAR10-DVS, respectively, over  state-of-the-art UDA algorithm. The two novel ideas that we proposed, i.e., contrastive learning loss and uncorrelated condition, lead  an additional 2.0\% and 3.7\% performance boosts, respectively, which puts in a competitive range with supervised learning methods. Specifically, we obtain a performance better than the best supervised model TDBN Resnet19\cite{https://doi.org/10.48550/arxiv.2203.06145} on the CIFAR10-DVS benchmark, despite the fact that TDBN is designed specifically to work well on event data.  This is a significant observation for practical purposes because it demonstrates that UDA can be used to train models to preform event tasks without data annotation. Please also note that E2VID and VID2E methods produce videos and use videos as a source domain, whereas we use images that makes our approach computationally less expensive.

\begin{table}[h]
\begin{center}
\small
\begin{tabular}{ c | c | c| c }      
\hline
Method&UDA&ResNet&Test Acc  \\  
\hline
 E2VID\cite{https://doi.org/10.48550/arxiv.1906.07165}&\checkmark & 34 & 82.1 \\ 
 VID2E\cite{https://doi.org/10.48550/arxiv.2110.10505}&\checkmark & 34& 80.7 \\  
 BG\cite{Messikommer20ral} & \checkmark&18 & 84.8  \\
 Baseline&\checkmark & 18 & 88.6  \\ 
 \textbf{DAEC$^2$}&\checkmark &  18& \textbf{90.6}  \\ 

 \hline
 E2VID\cite{https://doi.org/10.48550/arxiv.1906.07165}& $\times$ & 34 & 86.6  \\ 
 \textbf{VID2E}\cite{https://doi.org/10.48550/arxiv.2110.10505}&$\times$ &34 & \textbf{90.6}  \\ 
 EST\cite{https://doi.org/10.48550/arxiv.1904.08245}& $\times$& 34 & 81.7  \\ 
 HATS\cite{https://doi.org/10.48550/arxiv.1803.07913}& $\times$& 34 & 64.2  \\

 TDBN\cite{https://doi.org/10.48550/arxiv.2203.06145}& $\times$& 19 & 78.6  \\ 
\hline

\end{tabular}
\end{center}
\vspace{-5mm}
\caption{Classification accuracies on the N-Caltech101 dataset. Bold font denotes the best method in each category.}
\label{tabl1}
\end{table}

\begin{table}[h]
\small
\begin{center}
\begin{tabular}{ c | c | c| c }
\hline
Method &UDA & ResNet& Test Acc \\  
\hline
 BG(Our Repro.) & \checkmark& 18&  76.3 \\
 Baseline&\checkmark &  18&  79.4 \\ 
 \textbf{DAEC$^2$}&\checkmark &  18& \textbf{83.1}  \\

 \hline
 DART\cite{https://doi.org/10.48550/arxiv.1710.10800}& $\times$& N/A & 65.8 \\
 Spike-based BP\cite{https://doi.org/10.48550/arxiv.2007.05785}& $\times$& N/A & 74.8 \\

 \textbf{TDBN}\cite{https://doi.org/10.48550/arxiv.2203.06145}& $\times$&  19& \textbf{78.0}  \\

\hline

\end{tabular}
\end{center}
\vspace{-5mm}
\caption{Classification accuracies on the  CIFA10-DVS dataset.  Bold font denotes the best method in each category. }
\label{tabl2}
\vspace{-3mm}
\end{table}

\subsection{Ablative Experiments}
We perform an ablative experiment to demonstrate that our major ideas are crucial for improved performance.  Table~\ref{tabl3} presents our ablative study experiments using N-Caltech101 dataset. We have included the four possible scenarios for using contrastive learning and uncorrelated conditioning for model adaptation. The baseline denotes using all loss functions, except the two corresponding loss terms.
\begin{table}[h]
\small
\begin{center}
\vspace{-4mm}

\begin{tabular}{ c |c} 
\hline
Condition&Validation Acc. \\ 
\hline
Baseline Without Event Attribute Encoder&80.7\\
Baseline&82.5\\
With Contrastive Learning Loss&83.6\\
With Uncorrelated Condition&83.7\\
With Both &84.4\\
\hline
\end{tabular}
\end{center}
\vspace{-5mm} 
\caption{Ablative experiments  on N-Caltech101 dataset.}
\label{tabl3}
\vspace{-2mm}
\end{table}

We observe that extracting how objects look like under event cameras is necessary. Each of the two novel ideas improves model performance. When we use both ideas, the performance improves compared to using the two ideas individually. These experiments demonstrate that both of our ideas are important for obtaining an optimal performance.

\subsection{Analytical Experiments}
\begin{figure}[h]
    \centering
    \vspace{-5mm}    
    \includegraphics[width=5cm, height=4cm]{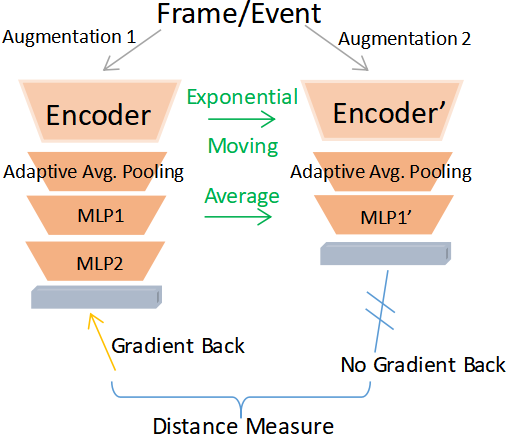}
    \vspace{-3mm}
    \caption{Momentum Encoder, labelled in $M_E$ in table \ref{tabl4}, if under average pooling, then there is no MLP1 and MLP1'.}
    \label{fig5}
\end{figure}

In this section, we provide empirical explorations about our algorithm to provide a better justification for DAEC$^2$.
\paragraph{Optimal Design:}  we used performance on the validation set to find the optimal design for our implementation. We explored four possibilities to implement the contrastive learning and uncorrelated conditioning constraints: 
\begin{itemize}
 \item  Directly impose constraints on feature maps:   We can use $\ell_1$ loss to align the two feature maps to enforce contrastive learning. However, it is not easy to implement uncorrelated conditioning on two feature maps. We just naively used negative $\ell_1$ loss.
\item Global average pooling on the feature maps: After global average pooling, we get a vector representation. For contrastive learning loss, we tried both $\ell_2$ distance and cosine similarity, i.e., either minimize $\ell_2$ distance or maximize the cosine similarity (in \ref{tabl4}, it is $-\cos$ because the loss is the negative of cosine similarity). For uncorrelated conditioning, the natural choice is cosine similarity. When its value is zero, it means the two vectors are uncorrelated. This means that the loss for this constraint is the absolute value of cosine similarity.
\item Adding MLP after global  pooling then computing distances: SimCLR\cite{https://doi.org/10.48550/arxiv.2002.05709} showed MLP can help. The constraints are implemented similarly to the previous case.

\item Inspired by BYOL~\cite{https://doi.org/10.48550/arxiv.2006.07733}, we also explored using momentum encoders and momentum MLP. As visualized in Figure~\ref{fig5}, the idea is when   we have two encoders, only one of them gets updated when gradient back propagates and the other one simply copies the parameters  using an exponential moving average. Momentum MLP is used when we  have one encoder but two MLPs. One MLP gets updated with back propagation and the other one simply copies the parameters using exponential moving average. BYOL\cite{https://doi.org/10.48550/arxiv.2006.07733} also shows that negative samples are not necessary, thus we do not use negative samples. Eliminating negative sampling allowed us to implement this condition with our limited computing resources.

\end{itemize}

\begin{table}
\begin{center}

\small
\begin{tabular}{ c |c| c c|c c} 

\multicolumn{6}{ c }{Contrastive Learning Loss Alone} \\
\hline
&  No Proj.&\multicolumn{2}{ c| }{Avg. Pl.} & \multicolumn{2}{ c }{MLP} \\
\hline
Loss& \(\ell_1\)&\(-\cos\)&\(\ell_2\)&\(-\cos\)&\(\ell_2\) \\
\hline
Val. Acc&83.5&\textbf{83.6}&82.4&83.1&82.7 \\
\hline
\end{tabular}
\end{center}

\begin{center}
\small
\begin{tabular}{ c |c| c |c} 

\multicolumn{4}{ c }{Uncorrelated Condition Alone} \\
\hline
&  No Proj.& Avg. Pl.&  MLP \\
\hline
Loss& \(-\ell_1\)&\(|\cos|\)&\(|\cos|\)\\
\hline
Val. Acc&82.2&\textbf{83.7}&82.6 \\
\hline
\end{tabular}
\end{center}

\begin{center}
\small
\begin{tabular}{ c |c |c |c |c} 

\multicolumn{1}{ c }{Val. Acc}&\multicolumn{1}{ c }{} &\multicolumn{1}{ c }{Contrastive  } &\multicolumn{1}{ c }{Uncorrelated } &\\
\hline
83.5&\multirow{1}{*}{No Proj.}& \(\ell_1\)&\(-\ell_1\)&Imp.\\
\hline
\textbf{84.4}&\multirow{4}{*}{Avg. Pl.}& \(-\cos\)&\multirow{4}{*}{\(|\cos|\)}&\\
83.6&& \(\ell_2\)&&\\
82.7&& \(-\cos\)&&$M_E$\\
82.4&& \(\ell_2\)&&$M_E$\\
\hline
82.2&\multirow{8}{*}{MLP}& \(-\cos\)&\multirow{8}{*}{\(|\cos|\)}&\\
84.0&& \(\ell_2\)&&\\
82.9&& \(-\cos\)&&$M_E$\\
83.3&& \(\ell_2\)&&$M_E$\\
84.2&& \(-\cos\)&&$M_M$\\
82.7&& \(\ell_2\)&&$M_M$\\
82.8&& \(-\cos\)& &S\\
83.5&& \(-\cos\)& &$M_M$S\\
\hline
\end{tabular}
\end{center}
\vspace{-5mm}
\caption{Optimal design search by validation accuracy on N-Caltech101 dataset for various implementations.}
\label{tabl4}
\vspace{-5mm}
\end{table}

\begin{figure*}[h]
    \centering
    
    \caption{UMap for data representations at the model output layers using the test data of CIFAR10 and CIFAR10-DVS. }
    \label{fig8}
    \vspace{-3mm}
    \includegraphics[width=17.3cm, height=4.2cm]{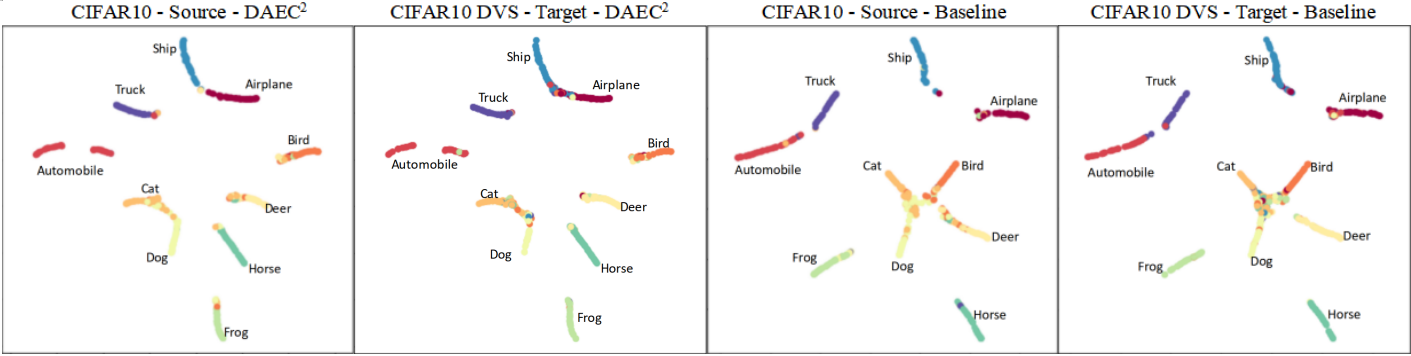}
    \vspace{-3mm}
\end{figure*}

\begin{SCfigure*}[0.5]
\caption{Images generated by three models. The input frame and event are draw from test set, and they are not paired. A random event from test set is paired with a frame, which is not shown. Green circles indicate that unnecessary lines are present in events generated by baseline model but should be discarded. 
These lines are boundaries of input frames, and are successfully discarded when contrastive learning loss and uncorrelated conditioning are introduced.}
\includegraphics[width=13cm, height=8cm]{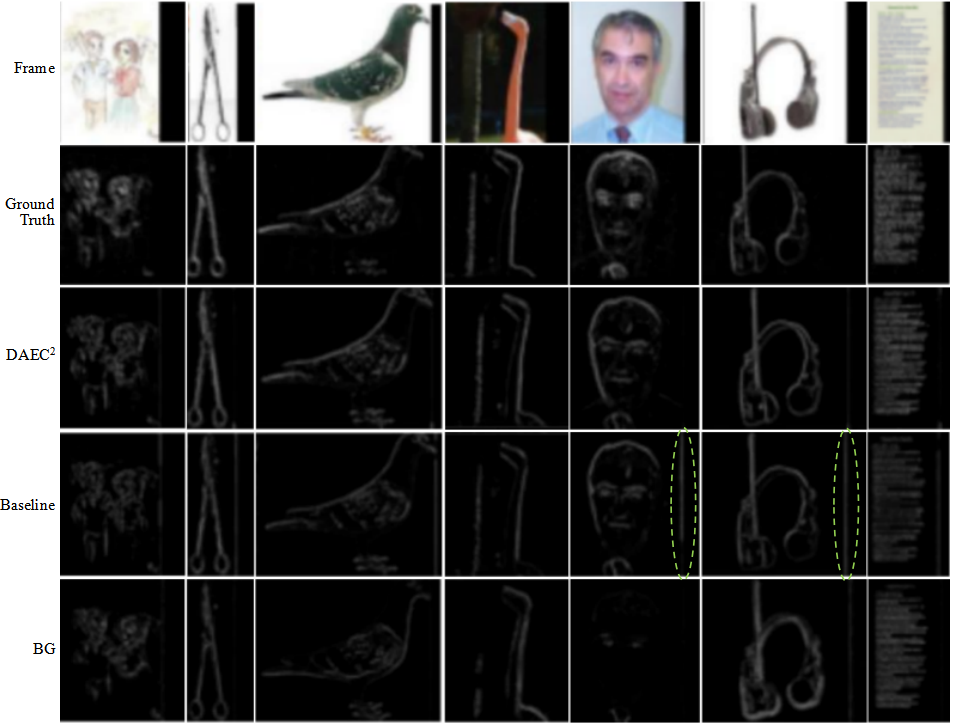}
\label{fig7}
\vspace{-4mm}
\end{SCfigure*}

Results for optimal design search are presented in Table \ref{tabl4}. 
In this table, $M_E$ means momentum encoders. If $M_E$ is listed under Avg. Pl., then there is no MLP1 and MLP1' in Figure~\ref{fig5}. $M_M$ stands for momentum MLP layers. For all MLPs, if not specified, the same MLP is used for event content and frame content, and a different MLP is used on event attribute. $S$ means using the same MLP for event content, frame content, and event attribute.

From the table, we observe that both contrastive learning loss and uncorrelated conditioning loss have positive contributions, irrespective of our design choices. However, we conclude that the optimal design for best validation performance is to use average pooling and apply cosine similarity measure to both constraints which leads to 1.1\% and 1.2\% performance increases,  respectively when used separately. Combining them leads to a 1.9\% performance increase. However, we see additional MLP layers, momentum encoders and momentum MLP do not lead to further improvement. We think the reason that an additional MLP is not helpful is that MLP gives the model more flexibility to get around with these two constraints. The purpose for these two constraints is to guide   encoders extract more informative features, but additional MLP weakens the guidance.  
Momentum encoders do not help for a similar reason: momentum encoders cannot provide a direct guidance to the encoders. In the case of BYOL\cite{https://doi.org/10.48550/arxiv.2006.07733}, momentum encoders are helpful because cosine similarity loss is its objective function, while we only need additional constraints as an extra help for the encoders. We also used a grid search on the weights of the two losses to tune them. We found out the best values are one for both.

\paragraph{2D visualization of features:}
In order to provide an intuitive justification, we used the UMap~\cite{mcinnes2018umap} visualization tool to reduce  the dimension of the   data representations in the latent embedding space  to two.
Figure \ref{fig8} shows the UMap~\cite{mcinnes2018umap} on the output of last layer of the classifier. All inputs are draw from test sets of CIFAR10 and CIFAR10-DVS.  
we observe that in both baseline and DAEC$^2$ have been effective and the two empirical distributions are aligned; however, class clusters are more separated and have less overlap when DAEC$^2$ is used. This observation demonstrates that our algorithm improves model generalizability and leads to training a model to domain shift in the input spaces.

\paragraph{Effect of DAEC$^2$ on the quality of generated images:}
Figure \ref{fig7} presents examples of generated fake events of our ground truth, DAEC$^2$, baseline,  and BG\cite{Messikommer20ral}. The inputs are drawn from test sets of Caltech101 and N-Caltech101, and they are not paired. A close visual inspection reveals that  our model leads to the best visual quality. Without our constraints, the model cannot identify and discard all useless features, i.e., the boundaries between frame images and the padding  (examples shown by green circles in Figure \ref{fig7}). We can see such unnecessary lines in almost every generated event by the baseline model. Contrastive learning loss and uncorrelated conditioning successfully identify these unnecessary features in input frames. Moreover, the events generated by DAEC$^2$ are clearer with fewer missing object boundaries. Thus, we can conclude that our proposed constraints help encoders to extract useful information and discard unnecessary information from the unpaired images.

\section{Conclusions}
\label{sec:Conclusions}
We introduced contrastive learning based on augmentation - invariant representation and uncorrelated conditioning as two novel ideas to perform UDA from frame-based data to event-based data, generally assumed to be a challenging UDA problem. Our experiments demonstrate both ideas are effective and lead to improved performance compared to the best existing UDA methods on two event-based benchmarks, leading to a performance comparable with supervised learning using annotated target domain data. Due to our limited computing resources, we could not perform other CV tasks like object detection, but we are confident that our ideas will work despite the task types. Uncorrelated conditioning on features extracted on the same input is an unprecedented ideas and its  effectiveness in UDA motivates its adoption in other tasks as our future work. Given our  empirical observations, uncorrelated conditioning can be used wherever we need to extract uncorrelated features from one same input, e.g., improving GAN data generation results or improving object detection.

{\small
\bibliographystyle{ieee_fullname}
\bibliography{egbib}
}
\end{document}